\documentclass[fleqn,10pt]{sadhana}
\usepackage[utf8]{inputenc}
\usepackage[T1]{fontenc}

\usepackage{subcaption}
\usepackage{graphicx}

\usepackage{xcolor, soul}
\sethlcolor{red}

\usepackage{wrapfig}
\usepackage{lscape}
\usepackage{rotating}
\usepackage{graphicx}
\usepackage{caption}
\usepackage{amsmath}
\usepackage{bm}
\usepackage{mathtools}
\usepackage{multirow,booktabs}
\usepackage{siunitx}
\usepackage{adjustbox}
\usepackage{graphicx}
\usepackage{float}

\usepackage{marginnote}
\usepackage{longtable}
\usepackage[sort&compress,square,numbers]{natbib}

\usepackage{algorithm}
\usepackage{algpseudocode}

\usepackage{nicematrix}
\usepackage{tikz}
\newcommand{\tikzmark}[2]{
    \tikz[overlay,remember picture,baseline] 
    \node[anchor=base] (#1) {$#2$};
}

\usepackage{breqn}

\usepackage{titlesec}

\begin{document}

\title{Enumeration of Spatial Manipulators by Using the Concept of Adjacency Matrix}

\author{Akkarapakam Suneesh Jacob\textsuperscript{1,*}, Bhaskar Dasgupta\textsuperscript{1}\and Rituparna Datta\textsuperscript{2}}
\affilOne{\textsuperscript{1} Indian Institute of Technology Kanpur, Kanpur, India\\}
\affilTwo{\textsuperscript{2} Capgemini Technology Services India Limited, Bengaluru, India\\}
\affilThree{e-mail: suneeshjacob@gmail.com; dasgupta@iitk.ac.in; rituparna.datta@capgemini.com}

\twocolumn[{

\maketitle

\begin{abstract}
This study is on the enumeration of spatial robotic manipulators, which is an essential basis for a companion study on dimensional synthesis, both of which together present a wider utility in manipulator synthesis. The enumeration of manipulators is done by using adjacency matrix concept. In this paper, a novel way of applying adjacency matrix to \emph{spatial} manipulators with four types of joints, namely revolute, prismatic, cylindrical and spherical joints, is presented. The limitations of the applicability of the concept to 3D manipulators are discussed. 1-DOF (Degree Of Freedom) manipulators of four links and 2-DOF, 3-DOF and 4-DOF manipulators of three links, four links and five links, are enumerated based on a set of conventions and some assumptions. Finally, 96 1-DOF manipulators of four links, 696 2-DOF manipulators of 5 links, 4 2-DOF manipulators of three links, 8 3-DOF manipulators of four links and 15 4-DOF manipulators of five links are presented.
\end{abstract}



}]


\markboth{Jacob, Dasgupta and Datta}{Enumeration of spatial manipulators by using the concept of Adjacency Matrix}

\section{Introduction and Literature Review}

Synthesis of a mechanism is conducted in many stages that include type synthesis, number synthesis, structural synthesis and dimensional synthesis. Type synthesis deals with designing the type of mechanism (such as types of joints, etc.) for a required task. Number synthesis deals with designing things such as the number of links, the number of joints and the number of Degrees Of Freedom (DOF) for the required task (also used for the study of determining the number of manipulators that can be enumerated for a given number of links and a given DOF). Dimensional synthesis deals with designing the dimensions of links, etc. Structural synthesis is the enumeration of all possible distinct mechanisms with the given number of links for a given DOF requirement.

One of the earliest studies on number synthesis of mechanisms in literature is done by Crossley \cite{crossley1964contribution} who introduced a simple algorithm to solve Gruebler's equation for constrained kinematic chains. In his paper, the mobility is chosen, and the Gruebler's equation is used to find how many binary links, ternary links, etc., would be needed to produce the chosen mobility. Correspondingly, the possible linkages would be enumerated and grouped. Franke \cite{davies1966structural} used a notation to transform a mechanism into a graph of nodes and edges. The nodes represent the non-binary links (ternary links, quaternary links, etc.), and each node bears the number of joints the corresponding link is connected to. The edges with multiple parallel lines (single line, double line, triple line, etc.) bear sequences of numbers that represent the number of connections between the corresponding two links. Graph theory is extensively used in the literature, in which the links of a mechanism are represented by vertices, joints by edges and joint connections by edge connections. Damir et al. \cite{vucina1991application} presented an application of graph theory to the kinematic synthesis of mechanisms. Manolescu \cite{manolescu1973method} presented a method based on Baranov Trusses to enumerate planar kinematic chains and mechanisms. Another famously used method in the literature for enumeration is the method of Assur groups \cite{asurgroup1}. Assur groups are formed by removing a link from Baranov Trusses. Assur group is a group of mechanisms that do not alter the DOF of that mechanism if added to a mechanism. Jinkui et al. \cite{jinkui1998systemics} presented Assur groups with multiple joints. Many more methods were used in literature to enumerate planar mechanisms. Mruthyunjaya \cite{mruthyunjaya2003kinematic} presented a review of methods for structural synthesis of planar mechanisms. Raicu \cite{raicu1974matrices} used the concept of adjacency matrix to capture the topological information. This matrix has all diagonal elements as zeroes and all off-diagonal elements as either zeroes or ones. The diagonal elements signify the links of the mechanism. The number 0 in each off-diagonal element signifies no connection between the two corresponding links. The number 1 signifies a joint that connects the two corresponding links. This type of representation of mechanisms can be useful in implementing the enumeration process on a computer, as permuting the off-diagonal elements of a matrix of a given size $n\times n$ would include the representation of all the possible mechanisms of $n$ links that are connected with joints. Mruthyunjaya et al. \cite{mruthyunjaya1979structural} proposed a generalised matrix notation to facilitate the representation and analysis of multiple-jointed chains. In their paper, multiple-joints are considered for planar mechanisms. Each of the off-diagonal elements in the matrix consists of a value that represents the number of links the joint is connected to. The concept of adjacency matrix for planar mechanisms is extensively used in the literature, especially for computerised enumeration of planar mechanisms. Li et al. \cite{compliantmech} presented application of adjacency matrix to compliant mechanisms. However, these were designed for planar mechanisms. Wenjian et al. \cite{yang2022structural} presented a review paper on the structural synthesis of planar mechanisms that covered the history of structural synthesis and its recent research progress. There are many studies in the literature \cite{simoni2009enumeration,huang2019structural,ding2012synthesis,pucheta2007automated,pozhbelko2015number,li2015assur,enumeration_application} that used various methods to present enumerated mechanisms. But even though these were very successful methods, unfortunately, these methods cannot be applied to spatial mechanisms. Moreover, these methods did not consider the distinction amongst mechanisms based on base and end-effector links. 

Amongst studies on spatial manipulators, enumeration of serial manipulators is straightforward but enumeration of parallel manipulators is non-trivial. Pierrot et al. \cite{pierrot1999h4} presented a new family of 4-DOF parallel robots by choosing different types of limbs of the end-effector plate. Dimiter et al. \cite{zlatanov2001family} also developed a family of new parallel manipulators of 4-DOF by analysing the inverted mechanism, i.e., considering the end-effector plate to be fixed and analysing the relative motion of the base link. An extension of Pierrot's work is proposed by Fang and Tsai \cite{fang2002structure} that includes development of a systematic method to enumerate 4-DOF and 5-DOF overconstrained parallel manipulators with identical serial limbs using screw theory. But this enumeration is limited to identical limb structures. Extending this, Hess-Coelho \cite{hess2007alternative} presented an alternative procedure for type synthesis of 2D and 3D parallel manipulators by employing asymmetric (non-identical) limbs. Li et al. \cite{spatial_new_metamorph} used a kind of adjacency matrix to describe metamorphic mechanisms with diagonal elements representing the joints and the off-diagonal elements representing the links. Bai et al. \cite{spatial_new_scaling} used adjacency matrix to describe scaling mechanisms in which off-diagonal elements represent links. Siying et al. \cite{long2022type} proposed a method to perform type synthesis of 6-DOF manipulators based on screw theory. Even though these studies on spatial manipulators were very useful, they did not use the advantages of the adjacency matrix concept. Qiang et al. \cite{5205134} used 12-bit string matrix representation of manipulators to perform structural synthesis of serial-parallel hybrid mechanisms. Extending this work, Zhang et al. \cite{zhang2011string} used a string-matrix based geometrical and topological representation of mechanisms. Their paper presents an extension of the 2D adjacency matrix concept to 3D by using 16 bits in each matrix element. Although the study opens up the application of the adjacency matrix concept to spatial manipulators, a proper study on the applicability of the concept to three dimensions is still needed for the justification of its usage and to understand its limitations for enumerating spatial manipulators. Moreover, there is no distinction based on base and end-effector links that is vital for enumerating \emph{robotic} manipulators. There exist studies \cite{kong_book} on enumeration of spatial parallel manipulators, however their scope is apparently limited to platform manipulators with limbs connected from the base link, which does not take into account many complex structures that are beyond platform manipulators (such as serial-parallel hybrid manipulators \cite{spm1,spm2}, etc.)

To summarise, many methods such as contracted graph methods, adjacency matrix methods and Assur groups exist in the literature and are used to enumerate 2D kinematic chains. Spatial manipulators have been enumerated using the adjacency matrix concept, but the direct application of adjacency matrix for 3D manipulators has not been studied, and moreover, the distinction based on base and end-effector links has not been made in the studies. The method presented in the present study extends the adjacency matrix application directly from 2D to 3D manipulators and takes into consideration the distinction based on base and end-effector links for four types of joints, namely revolute, prismatic, cylindrical and spherical.

An important issue in mechanism synthesis is isomorphism detection and elimination. Isomorphism detection is generally considered to be a difficult problem due to the number of permutations it involves for a large number of nodes and there are several studies on handling isomorphism \cite{isomorphism1,isomorphism2,pucheta2007automated}. However, for reduction in computational load, manipulators only up to 5 links are considered for enumeration. For 4-node and 5-node graph adjacency matrices, the permutations involved are 4! And 5! respectively. Moreover, in the current study, the permutations are required for links other than the first and the last links, and hence the total permutations involved for each adjacent matrix are just 2! and 3! for 4-link and 5-link robots, respectively. Hence, brute-force search is considered to detect isomorphism, which is basically to produce all possible isomorphic adjacency matrices of a given adjacency matrix and compare it with other adjacency matrices in the enumerated list, to check for any match.

\section{Analysis and Methodology}

\subsection{Adjacency matrix representation}

\begin{figure}[hbt!]
  \centering
  \includegraphics[width=\linewidth]{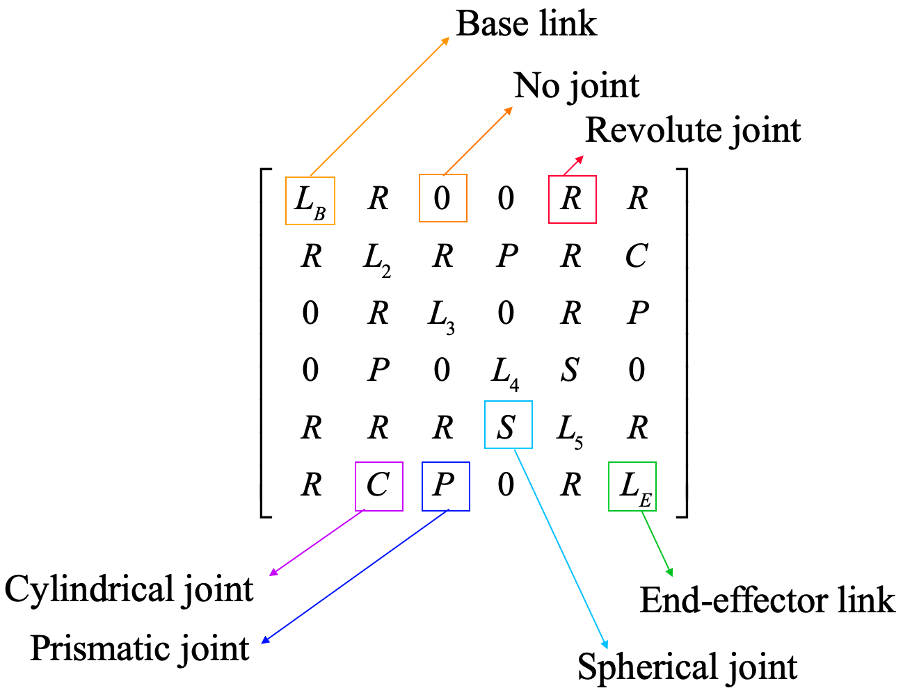}
  \caption{Typical adjacency matrix structure}
  \label{fig_adjmat}
\end{figure}

The adjacency matrix notation used to describe manipulators in this study is an $n\times n$ symmetric matrix where $n$ is the number of links of the manipulator. The diagonal elements represent the links. Moreover, the first diagonal element represents the base link and the last diagonal element represents the end-effector link. Each off-diagonal element represents the joint connecting the two links corresponding to its indices. A typical adjacency matrix structure is shown in Figure \ref{fig_adjmat}.

\subsection{On the applicability of adjacency matrix representation of spatial manipulators}

The concept of adjacency matrix representation for enumerating 2D manipulators is extended to 3D in this study. Regarding the compatibility of adjacency matrix representation with 3D manipulators, the main constraint of adjacency matrix representation is that it allows a maximum of one joint to be the connection between any two links. The joints considered in this study are Revolute, Prismatic, Cylindrical and Spherical joints. Additionally, in this study, it is assumed that only prismatic and revolute joints are capable of serving as actuators. Since this study considers four types of joints, there would be $\left(\frac{4^2-4}{2!}\right)+4=10$ possible cases of links sharing two joints, namely the revolute-revolute case, the revolute-prismatic case, the revolute-cylindrical case, the revolute-spherical case, the prismatic-prismatic case, the prismatic-cylindrical case, the prismatic-spherical case, the cylindrical-cylindrical case, the cylindrical-spherical case and the spherical-spherical case. In all the cases except the spherical-spherical case, the relative motion is either impossible or not guaranteed for arbitrary positions and orientations of axes of joints, and therefore these cases are omitted in the enumeration. In the spherical-spherical case, two links connected with two spherical joints at arbitrary positions can have relative motion. The kind of relative motion in such a case would be equivalent to the relative motion of two links connected with a single revolute joint, where the axis of the revolute joint would be the line passing through the two centres of the two spherical joints. The above cases show that two links connected by two joints (at arbitrary locations and orientations) can have motion only in the spherical-spherical case. Since the motion in this case is equivalent to the motion of two links connected by a single revolute joint, this case is omitted in the enumeration such that the entire possible enumerations that are not omitted would consist of at most one joint connecting any two links. Thus, such enumeration can be made by permutating the off-diagonal elements of an adjacency matrix.

\subsection{Methodology to enumerate manipulators}

\quad To enumerate $n$-link manipulators, the concept of adjacency matrix is used in this study. All the possible adjacency matrices of $n$ links with the four kinds of joints are generated using Python, and the criteria, listed below, are used to eliminate invalid and isomorphic adjacency matrices. The set of all the possible $n\times n$ adjacency matrices would capture all the possible $n$-link manipulators. Since an $n\times n$ adjacency matrix is symmetric with the diagonal elements representing the links, there would be $\frac{n^2-n}{2}$ independent places of the matrix that can be filled with joints.
Since each of these places can be either filled with one of the four joints or left empty, there would be $4+1$ possible types of connection between any two links. By filling the places with all the possible types of connection, $5^{\left(\frac{n^2-n}{2}\right)}$ distinct adjacency matrices can be formed, among which some would not qualify and some would be isomorphic.

The criteria listed below are used to eliminate invalid and isomorphic mechanisms.\newline
\label{criteria}

$\bullet$ \textbf{DOF should be greater than or equal to 1.} \newline
\quad \quad Since the enumerated adjacency matrices would have structures (including indeterminate ones) included, these are removed by identifying the DOF using the Kutzbach criterion.\newline

$\bullet$ \textbf{The mechanism should have at least one revolute joint or prismatic joint.} \newline
\quad Only revolute and prismatic joints are considered for actuation, and therefore at least one prismatic or revolute joint is needed in a mechanism.\newline

$\bullet$ \textbf{The sum of the numbers of prismatic and revolute joints should be greater than or equal to the DOF.} \newline
\quad Since the actuation is given through only prismatic and revolute joints, the sum of the numbers of prismatic and revolute joints must be greater than or equal to the DOF of the manipulator.\newline

$\bullet$ \textbf{The mechanism should not have any link that is entirely unconnected from all the other links.} \newline
\quad One possible adjacency matrix for a five-link manipulator is shown in Equation \eqref{eq:adj_mat_for_isolated_link_example} in which, the third link, i.e., $L_3$, has no connection with any other link and therefore is not part of the mechanism, and hence such mechanisms should be removed.\newline
\begin{equation}
\label{eq:adj_mat_for_isolated_link_example}
A=\left[\begin{matrix}L_{1}&O&O&R&P\\O&L_{2}&O&C&S\\O&O&L_{3}&O&O\\R&C&O&L_{4}&O\\P&S&O&O&L_{5}\end{matrix}\right]
\end{equation}

\begin{figure}[hbt!]
  \centering
  \includegraphics[width=\linewidth]{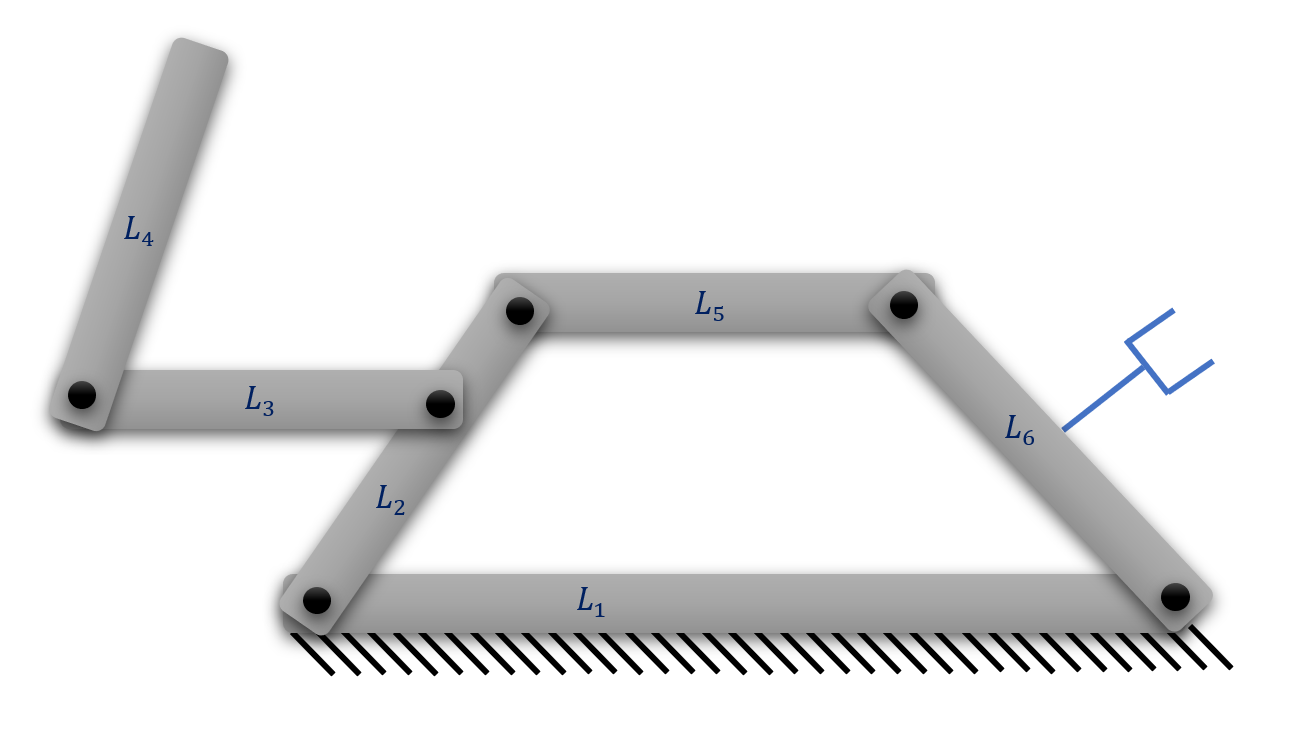}
  \caption{Manipulator with a non-contributing open-chain}
  \label{fig_illustration_1}
\end{figure}

$\bullet$ \textbf{The mechanism should not have open-chains that do not have a connection from the base-link to the end-effector link.} \newline
\quad In figure \ref{fig_illustration_1}, the mechanism has the open-chain of links 3 and 4 that do not have a connection from the base-link to the end-effector link. Such mechanisms are removed.\newline

\begin{figure}[hbt!]
  \centering
  \includegraphics[width=\linewidth]{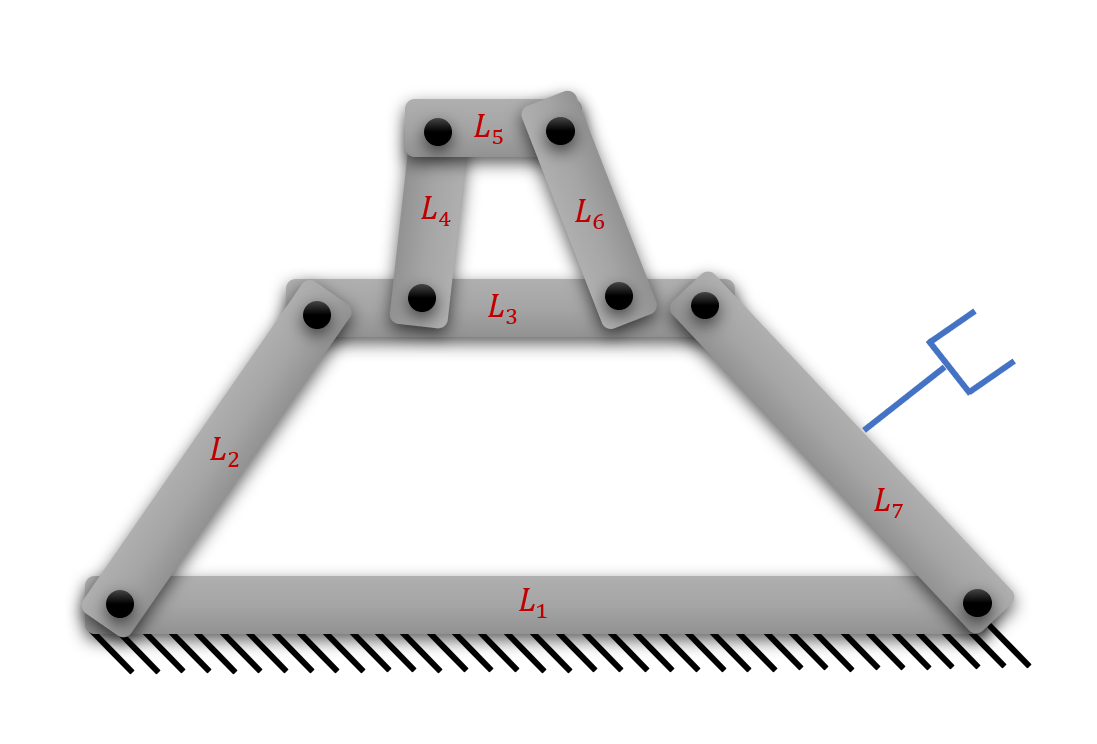}
  \caption{Manipulator with a non-contributing loop}
  \label{fig_illustration_2}
\end{figure}

$\bullet$ \textbf{The mechanism should not have non-contributing loops.} \newline
\quad In figure \ref{fig_illustration_2}, the mechanism has the $L_3-L_4-L_5-L_6$ loop non-contributing and therefore cannot contribute, either actively or passively, to the transmission of velocity to the end-effector. Hence such mechanisms are to be removed.\newline

$\bullet$ \textbf{The mechanisms should not be isomorphic.} \newline
\quad Since two isomorphic adjacency matrices represent the same mechanism, one of them needs to be removed. In this study, the first link represents the base link, and the last link represents the end-effector link. Therefore, isomorphism in this context is defined to be the mechanism-repetition for a unique set of base and end-effector links.\newline

$\bullet$ \textbf{The end-effector should not have a connection with only two spherical joints.} \newline
\quad If the end-effector of a manipulator is connected to the rest of the mechanism through two spherical joints and no other joint, in which case the end-effector is connected to one link with a spherical joint and to another link with the other spherical joint, then the relative velocity of the end-effector about the axis passing through the two spherical joints, cannot be controlled by the actuators. Hence, such mechanisms are to be removed. \newline

$\bullet$ \textbf{The actuation should not be locked within a sub-mechanism of the main mechanism.} \newline
\quad If an actuator-joint gets permanently locked within a sub-mechanism of the main mechanism for all its configurations by virtue of the kinds of joints connected to the links, then the actuation cannot be feasible from that joint. Hence, if a mechanism has the sum of non-locked revolute and non-locked prismatic joints less than the required DOF, then such matrices are removed.\newline

\begin{figure}[hbt!]
  \centering
  \includegraphics[width=\linewidth]{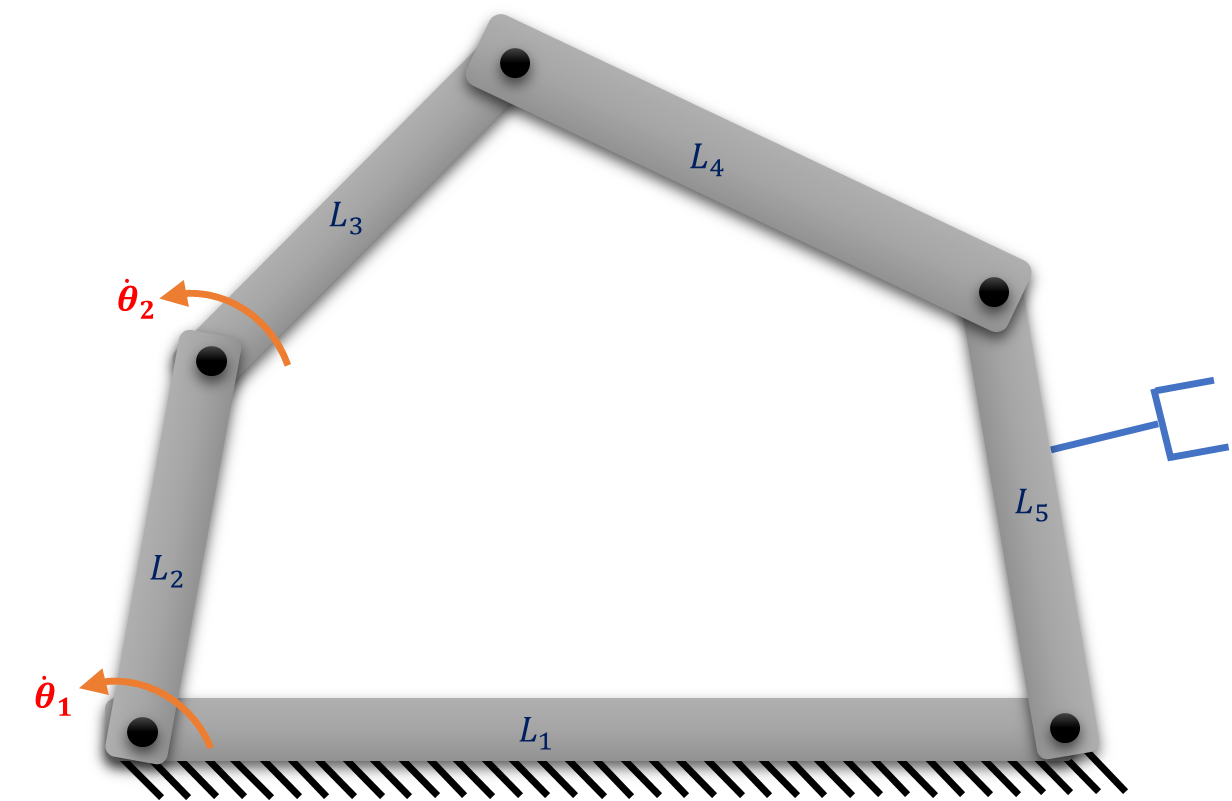}
  \caption{Manipulator having two actuators and one independent component of end-effector velocity}
  \label{fig_illustration_3}
\end{figure}

$\bullet$ \textbf{The number of independent directions of the final velocity of the end-effector should be equal to the number of independent actuations.} \newline
\quad In figure \ref{fig_illustration_3}, assuming all the links are connected with revolute pairs, even if both the actuating velocities $\dot{\theta}_1$ and $\dot{\theta}_2$ are independently provided, the end-effector can have only one independent component of velocity, and hence such manipulators are to be removed.\newline

\begin{figure}[hbt!]
  \centering
  \includegraphics[width=\linewidth]{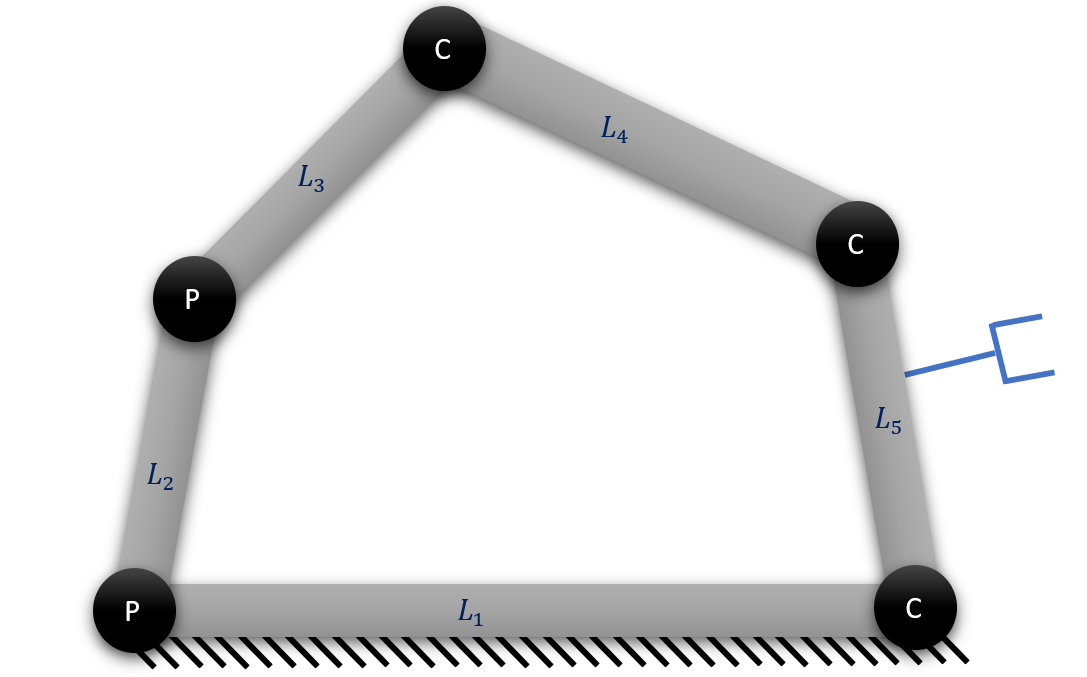}
  \caption{Schematic diagram of a manipulator having less than three joints of revolute motion in a loop}
  \label{fig_illustration_4}
\end{figure}

$\bullet$ \textbf{The number of arbitrarily positioned and oriented joints that contribute independent angular motions in a loop should be more than 3 for the motion to be possible in any of the joints.} \newline
\quad In the figure \ref{fig_illustration_4}, the loop $L_1-L_2-L_3-L_4-L_5-L_1$ has only three joints that can facilitate relative angular motions between the links connected to them, namely the cylindrical joint connecting $L_3 \;\text{and}\; L_4$, the cylindrical joint connecting $L_4\;\text{and}\;L_5$ and the cylindrical joint connecting $L_5,L_1$. Thus, in order for the joints to facilitate angular motion, the resultant angular velocity produced by the joint connecting $L_3\;\text{and}\;L_4$ and the joint connecting $L_4\;\text{and}\;L_5$ should be about the same axis as the joint connecting $L_5\;\text{and}\;L_1$, which is a restricted case. Assuming that the positions and the orientations of the joints are arbitrary, none of the cylindrical joints would produce angular motion but can produce only linear motion and therefore all the rotations of the cylindrical joints would be locked. Hence, at least four independent angular motions are needed for arbitrarily located and oriented joints to accommodate angular motion in a loop. \newline

\begin{figure}[hbt!]
    \centering
    \includegraphics[width=\linewidth]{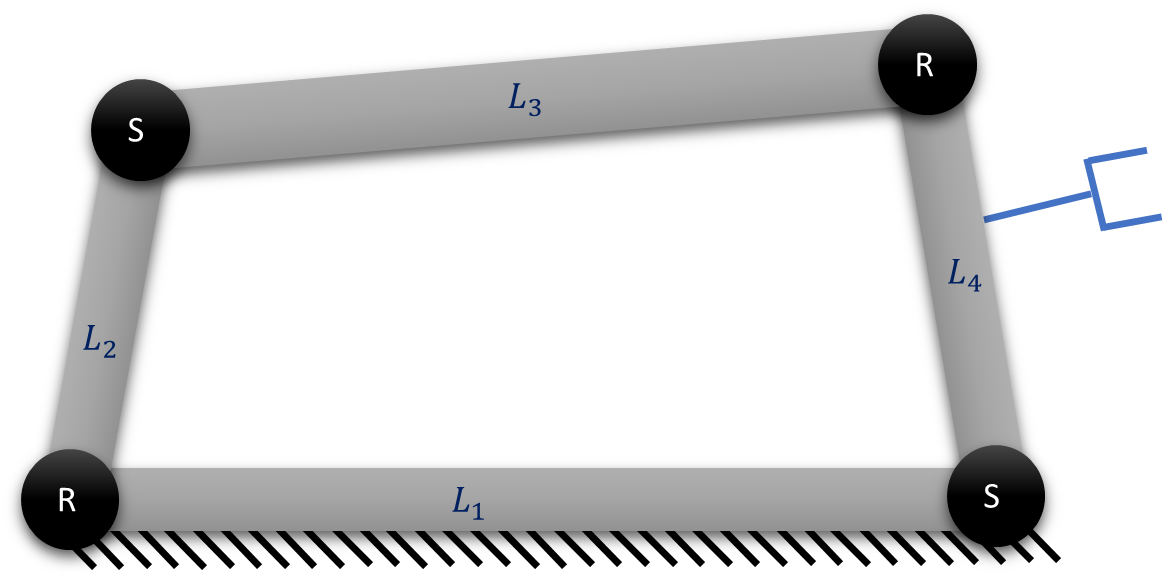}
    \caption{Schematic diagram of a manipulator having its end-effector uncontrollable by its actuator}
    \label{fig_illustration_7}
  \end{figure}

\begin{figure}[hbt!]
    \centering
    \includegraphics[width=\linewidth]{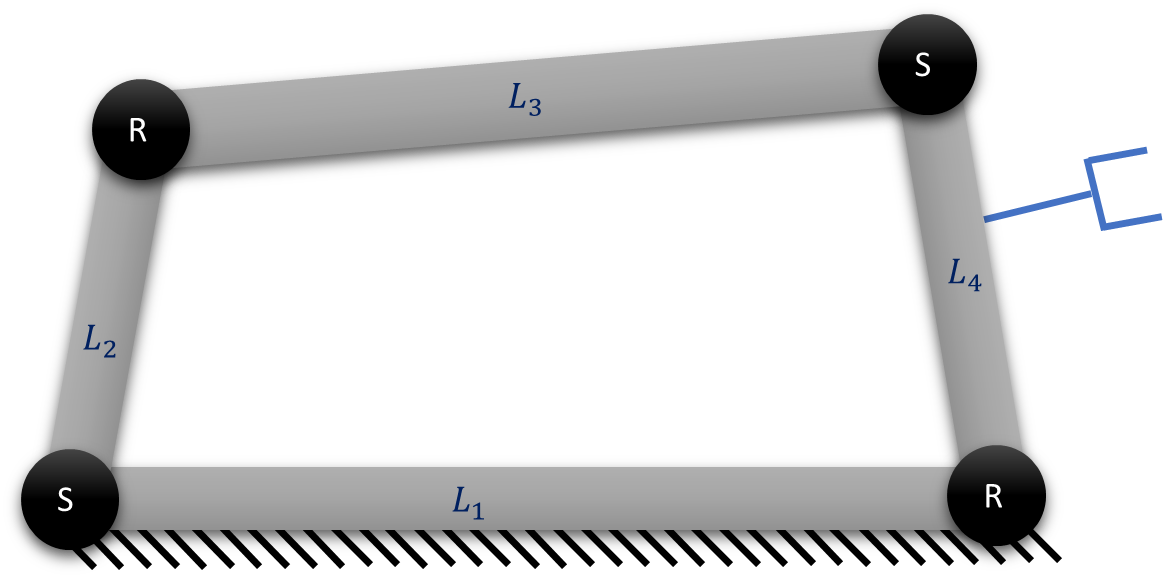}
    \caption{Schematic diagram of a manipulator having a part (of set of links) uncontrollable by its actuator}
    \label{fig_illustration_8}
  \end{figure}

$\bullet$ \textbf{The mechanism should not have parts of it uncontrollable by the actuators except as superfluous DOF mechanisms.} \newline
\quad If the manipulator has any two of its complementary parts connected by two spherical joints and no other joint, and if the base link lies on one part whilst the end-effector link lies on the other part, then the rotation of the end-effector link along the axis passing through centres of the two spherical joints cannot be controlled by the actuators. An example of it is the manipulator shown by the schematic diagram in \ref{fig_illustration_7}, in which the rotation of the part consisting of links 3 and 4 cannot be controlled by the actuator along the axis passing through the centres of two spherical joints. Hence, such mechanisms are to be removed. Furthermore, if both the base link and the end-effector link lie on the same part and if the other part comprises more than one link, the manipulator cannot control the rotation of that part about the axis passing through the centres of the two spherical joints by the actuators. An example of such a case is shown in \ref{fig_illustration_8}, where the angular velocity of the part of manipulator consisting of links 2 and 3 about the axis passing through the centres of the two spherical joints, cannot be controlled by the actuator of the manipulator. Hence, such mechanisms are also removed. Even though the latter case does not affect the velocity of the end-effector, such manipulators are removed with the convention that any moving parts of mechanism that cannot be controlled by the actuators are removed except if the uncontrollable part of mechanism consists of just one link. The inclusion of such manipulators is suggested as future scope.\newline

$\bullet$ \textbf{If all the velocities of the end-effector are of purely linear motion then the DOF of the manipulator cannot exceed 3.} \newline
\quad If all the joints in the manipulator are prismatic then the end-effector cannot have angular motion but can only have linear motion. Since the 3D space allows only three independent linear components and three angular components of velocity of the end-effector, the DOF that can be attained by mere linear motion can at most be 3, as any additional linear actuation would cause a motion that is within the three linear components of the end-effector's velocity. Hence, if a manipulator has only prismatic type of joints but has its DOF more than 3, then such manipulator is omitted.\newline

\section{Algorithms}

This section discusses the algorithms used to produce the results shown in the next section. Firstly, all possible adjacency matrices are to be enumerated. In an $n\times n$ adjacency matrix, there would be $n$ diagonal elements and $n^2-n$ off-diagonal elements. Since adjacency matrices are symmetric matrices, there would be only $\frac{n^2-n}{2}$ independent entries in the off-diagonal elements. As an example, for $n=4$ there would be $\frac{4^2-4}{2}=6$ independent entries ($x_1, x_2, x_3, x_4, x_5$ and $x_6$) as shown below.

$$\begin{bmatrix}L_1 & x_1 & x_2 & x_3 \\
x_1 & L_2 & x_4 & x_5 \\
x_2 & x_4 & L_3 & x_6 \\
x_3 & x_5 & x_6 & L_4
\end{bmatrix}$$

The candidate values for each $x_i$ are $O, R, P, C$ and $S$\footnote{$O, R, P, C$ and $S$ represent the absence of a joint, a revolute joint, a prismatic joint, a cylindrical joint and a spherical joint, respectively. In the actual programming, these values were conventionally considered to be $0, 1, 2, 3$ and $4$, respectively, and each diagonal element was filled with the value $9$.}. The array $[x_1, x_2, x_3, x_4, x_5, x_6]$ with each $x_i$ having a potential value from one of the five values ($O, R, P, C$ and $S$) can be permuted with a total of $5^{6}=15625$ permutations. After permuting, these $x_i$ values can be put in the above matrix to form 15625 distinct matrices. Among these, there would be invalid and isomorphic matrices which need to be removed.

In order to implement the criteria discussed in the previous section, 12 conditions are outlined below from algorithmic point of view.

\subsection{Conditions 1 and 2}
The first condition is that the sum of prismatic and revolute joints should be greater than 1. The second condition is that the DOF computed by Kutzbach criterion should be greater than or equal to 1. These two conditions can be computed from the number of revolute joints ($n_r$), the number of prismatic joints ($n_p$), the number of cylindrical joints ($n_c$) and the number of spherical joints ($n_s$) together. The formula to find out the DOF based on Kutzbach criterion is $6(n-1)-5(n_r+n_p)-4n_c-3n_s$, where $n$ is the number of links. A pseudocode for these conditions is shown in Algorithm \ref{alg:conditions12}.

\begin{algorithm}
\caption{Pseudocode for conditions 1 and 2}\label{alg:conditions12}
\begin{algorithmic}[1]
\Require The adjacency matrix $M$.
\Ensure The boolean values $c_1$ and $c_2$.
\State $n_r$ = number of revolute joints in $M$.\;
\State $n_p$ = number of prismatic joints in $M$.\;
\If{$n_r+n_p<1$}
\State $c_1=false$\;
\EndIf
\State $D = $ DOF of $M$ based on Kutzbach criterion.\;
\If{$D < 1$}
\State $c_2=false$\;
\EndIf
\State Return $c_1$ and $c_2$.\;
\end{algorithmic}
\end{algorithm}

\subsection{Condition 3}
The third condition is to address degeneracy, i.e., the permanent locking of a sub-mechanism within the main mechanism of the robot. This is addressed by finding out all possible loops and analysing each loop to check whether the loop is able to exhibit reciprocating motion or not. For each loop, two quantities $q_r$ and $q_p$ are computed, where $q_r$ denotes the total number of revolute components facilitated by the joints in the loop, and $q_p$ represents the total number of prismatic components facilitated by the joints. A revolute joint facilitates one revolute component, a prismatic joint facilitates one prismatic component, a cylindrical joint facilitates a revolute component and a prismatic component, and a spherical joint facilitates three revolute components. In addition to this, if two spherical joints exist in a loop, then $q_r$ needs to be reduced by 1, because the revolution of the part (of links) that connects these two joints (with respect to the other part) would be a motion decoupled from the reciprocating motion of the loop. If more than two spherical joints exist then for each combination of two spherical joints there would be one such decoupled motion and hence $n_r$ needs to be reduced by ${}^{m_s}C_2$ where $m_s$ is the number of spherical joints. The sum of the two quantities $q_r$ and $q_p$ is denoted by $q_t$. The formulas for $q_r$, $q_p$ and $q_t$ are shown in Equations \eqref{eq:1}, \eqref{eq:2} and \eqref{eq:3} respectively, where $m_r, m_p$ and $m_c$ denote the numbers of revolute joits, prismatic joints and cylindrical joints of the loop, respectively.

\begin{align}
q_r &= m_r + m_c + 3m_s
\label{eq:1} \\
q_p &= m_p + m_c
\label{eq:2} \\
q_t &= q_r + q_p
\label{eq:3}
\end{align}

For motion to happen, $q_t$ needs to be more than 6. In special case where $0\leq q_r < 4$, the loop can still reciprocate if $q_p$ is more than 4, because four prismatic joints are sufficient to reciprocate motion in a loop. If the reciprocation is found to be not possible for a mechanism then that shows that there is degeneracy in the mechanism, and such corresponding adjacency matrices are considered to be invalid in this study. A pseudocode for the same is shown in Algorithm \ref{alg:condition3}.

\begin{algorithm}
\caption{Pseudocode for condition 3}\label{alg:condition3}
\begin{algorithmic}[1]
\Require Adjacency matrix $M$.
\Ensure The boolean value $c_3$.
\State Assume $c_3$ is true by default.\;
\State Get the list of all the loops from matrix $M$ as $A$.
\For{$i$ through $A$}
\State Get all the joints contained in the loop $i$ as $J$.\;
\State $m_r = $ number of revolute joints in $J$.\;
\State $m_p = $ number of prismatic joints in $J$.\;
\State $m_c = $ number of cylindrical joints in $J$.\;
\State $m_s = $ number of spherical joints in $J$.\;
\State $q_r = m_r + m_c +3m_s - {}^{m_s}C_2$.\;
\State $q_p = m_p + m_c$.\;
\State $q_t = q_r + q_p$.\;
\If{$q_r < 4$}
\If{$q_p < 4$}
\State $c_3 = false$.\;
\State Break the loop.\;
\EndIf
\If{$q_t < 7$}
\State $c_3 = false$.\;
\State Break the loop
\EndIf
\EndIf
\EndFor
\State Return $c_3$.\;
\end{algorithmic}
\end{algorithm}

\subsection{Conditions 4, 5, 6, 7 and 8}

The conditions 4, 5, 6, 7 and 8 are grouped together because they all require consideration of splitting of mechanism into two parts. The fourth condition is as follows: if the manipulator is split into two parts, and if the corresponding coupling matrix has only one joint then the base link and the end-effector link should not lie in the same part. This ensures that the mechanism does not have non-contributing open-chains. As an example, in Figure \ref{fig_illustration_1}, a mechanism with non-contributing open-chains is shown. The corresponding adjacency matrix $M$ would be as shown in Equation \eqref{eq:adj_mat_for_fig2}.

\begin{equation}
\label{eq:adj_mat_for_fig2}
M=\left[\begin{matrix}L_{1}&R&O&O&O&R\\R&L_{2}&R&O&R&O\\O&R&L_{3}&R&O&O\\O&O&R&L_{4}&O&O\\O&R&O&O&L_{5}&R\\R&O&O&O&R&L_6\end{matrix}\right]
\end{equation}

Since it is a 6-link manipulator, there would be ${}^{6}C_1+{}^{6}C_2+{}^{6}C_3=41$ possible ways of splitting the links into two parts. One such possible way is splitting the links into $p_1=\{1,2,5,6\}$ and $p_2=\{3,4\}$. By preserving the connections between links, the embedded matrix $\widetilde{M}$ can be formed as shown in Equation \eqref{eq:embedded_matrix}, from which the coupling matrix $C$ can be extracted as shown in Equation \eqref{eq:coupling_matrix}.

\begin{equation}
\label{eq:embedded_matrix}
    \widetilde{M} = \begin{pNiceMatrix}[margin]
    \tikzmark{c1_1}{L_3} & R & \Block[draw,fill=blue!15,rounded-corners]{2-4}{} O & R & O & O \\
    R & \tikzmark{c1_2}{L_4} & O & O & O & O \\
    O & O & \tikzmark{c2_2}{L_1} & R & O & R \\
    R & O & R & \tikzmark{c2_3}{L_2} & R & O \\
    O & O & O & R & \tikzmark{c2_4}{L_5} & R \\
    O & O & R & O & R & \tikzmark{c2_5}{L_6}
    \end{pNiceMatrix}
\end{equation}
\begin{tikzpicture}[overlay,remember picture]
     \draw[opacity=.2,line width=3mm,line cap=round] (c1_1.center) -- (c1_2.center);
     \draw[opacity=.2,line width=3mm,line cap=round] (c2_2.center) -- (c2_5.center);
\end{tikzpicture}
\begin{equation}
\label{eq:coupling_matrix}
    C = \begin{bmatrix}
    O & R & O & O \\
    O & O & O & O
    \end{bmatrix}
\end{equation}

Now, it can be observed that $C$ has only one joint and that both the base link ($L_1$) and the end-effector link ($L_6$) lie in the same part ($p_1$ in this case). This shows that $p_2$ is not contributing in transmitting motion from the base link to the end-effector link. Hence, such matrices are to be removed. This is considered to be the fourth condition in this study. On the other hand, if base link ($L_1$) happens to lie in one part and the end-effector link ($L_6$) lies in the other part, then the controllability of the end-effector's motion needs to be powered through this joint. Since only revolute and prismatic joints are considered to be capable of being actuators in this study, the joint that connects these two parts should be either a revolute joint or a prismatic joint. Hence, the matrices for which such a joint happens to be either cylindrical or spherical, are considered to be invalid. This is considered to be the fifth condition in this study. And an example of it is a serial manipulator with a cylindrical or spherical joint(s).

When the number of joints in the coupling matrix happens to be more than one, if all the joints happen to be connected to one link of a part and both the base and the end-effector links lie on that part, then that apparently shows that the other part is non-contributing to the motion of the end-effector link and hence such cases are considered invalid. This is considered to be the sixth condition in this study. An example of this can be the manipulator shown in Figure \ref{fig_illustration_2}. The adjacency matrix for this example is shown below.

\begin{equation}
\label{eq:main_matrix2}
M=\left[\begin{matrix}L_{1}&R&O&O&O&O&R\\R&L_{2}&R&O&O&O&O\\O&R&L_{3}&R&O&R&R\\O&O&R&L_{4}&R&O&O\\O&O&O&R&L_{5}&R&O\\O&O&R&O&R&L_6&O\\R&O&R&O&O&O&L_7\end{matrix}\right]
\end{equation}
\begin{equation}
\label{eq:embedded_matrix2}
    \widetilde{M} = \begin{pNiceMatrix}[margin]
    \tikzmark{c1_1}{L_1} & R & R & \Block[draw,fill=blue!15,rounded-corners]{3-4}{} O & O & O & O \\
    R & \tikzmark{c1_2}{L_2} & O & R & O & O & O \\
    R & O & \tikzmark{c1_3}{L_7} & R & O & O & O \\
    O & R & R & \tikzmark{c2_1}{L_3} & R & O & R \\
    O & O & O & R & \tikzmark{c2_2}{L_4} & R & O \\
    O & O & O & O & R & \tikzmark{c2_3}{L_5} & R \\
    O & O & O & R & O & R & \tikzmark{c2_4}{L_6}
    \end{pNiceMatrix}
\end{equation}
\begin{tikzpicture}[overlay,remember picture]
     \draw[opacity=.2,line width=3mm,line cap=round] (c1_1.center) -- (c1_3.center);
     \draw[opacity=.2,line width=3mm,line cap=round] (c2_1.center) -- (c2_4.center);
\end{tikzpicture}
\begin{equation}
\label{eq:coupling_matrix2}
    C = \begin{bmatrix}
    O & O & O & O \\
    R & O & O & O \\
    R & O & O & O
    \end{bmatrix}
\end{equation}

The coupling matrix corresponding to $p_1=\{1,2,7\}$ and $p_2=\{3,4,5,6\}$ is shown in Equation \eqref{eq:coupling_matrix2}. From the coupling matrix, it can be seen that all the joints are connected to the third link (of $p_2$). Furthermore, it can be seen that the base link ($L_1$) and the end-effector link ($L_7$) lie in the same part ($p_1$ in this case). Hence it fails to meet the sixth condition and hence stands invalid.

In particular, when the coupling matrix happens to have exactly two joints with both of being spherical joints, if it is not the case that all the joints happen to be connected to one link of a part, then a case like Figure \ref{fig_illustration_7} or Figure \ref{fig_illustration_8} would occur, both of which are considered to be invalid for this study. This is the seventh condition. As an example, Figure \ref{fig_illustration_8} is represented by the adjacency matrix shown in Equation \eqref{eq:main_matrix3}. For a split of $p_1=\{1,4\}$ and $p_2=\{2,3\}$, the corresponding coupling matrix is as shown in Equation \eqref{eq:coupling_matrix3}. It can be seen that the two joints in the coupling matrix are not connected to the same link but rather connected to different links (one spherical joint is connected to links 1 and 2, and the other spherical joint is connected to the links 3 and 4 and hence there is no common link that is connected to the two joints), and hence such matrices are considered invalid for this study.

\begin{equation}
\label{eq:main_matrix3}
M=\left[\begin{matrix}L_{1} & S & O & R\\S&L_{2}&R&O\\O&R&L_{3}&S\\R&O&S&L_{4}\end{matrix}\right]
\end{equation}
\begin{equation}
\label{eq:embedded_matrix3}
    \widetilde{M} = \begin{pNiceMatrix}[margin]
    \tikzmark{c1_1}{L_1} & R & \Block[draw,fill=blue!15,rounded-corners]{2-2}{} S & O \\
    R & \tikzmark{c1_2}{L_4} & O & S \\
    S & O & \tikzmark{c2_1}{L_2} & R \\
    O & S & R & \tikzmark{c2_2}{L_3}
    \end{pNiceMatrix}
\end{equation}

\begin{tikzpicture}[overlay,remember picture]
     \draw[opacity=.2,line width=3mm,line cap=round] (c1_1.center) -- (c1_2.center);
     \draw[opacity=.2,line width=3mm,line cap=round] (c2_1.center) -- (c2_2.center);
\end{tikzpicture}
\begin{equation}
\label{eq:coupling_matrix3}
    C = \begin{bmatrix}
    S & O \\
    O & S
    \end{bmatrix}
\end{equation}

Likewise, the coupling matrix corresponding to the split of $p_1=\{1,2\}$ and $p_2=\{3,4\}$ is shown in Equation \eqref{eq:coupling_matrix4}, where it is again not the case that all the joints are having a common link. Hence such matrices are also considered to be invalid.

\begin{equation}
\label{eq:embedded_matrix4}
    \widetilde{M} = \begin{pNiceMatrix}[margin]
    \tikzmark{c1_1}{L_1} & R & \Block[draw,fill=blue!15,rounded-corners]{2-2}{} O & S \\
    R & \tikzmark{c1_2}{L_2} & S & O \\
    O & S & \tikzmark{c2_1}{L_3} & R \\
    S & O & R & \tikzmark{c2_2}{L_4}
    \end{pNiceMatrix}
\end{equation}

\begin{tikzpicture}[overlay,remember picture]
     \draw[opacity=.2,line width=3mm,line cap=round] (c1_1.center) -- (c1_2.center);
     \draw[opacity=.2,line width=3mm,line cap=round] (c2_1.center) -- (c2_2.center);
\end{tikzpicture}
\begin{equation}
\label{eq:coupling_matrix4}
    C = \begin{bmatrix}
    O & S \\
    S & O
    \end{bmatrix}
\end{equation}

\begin{figure}[hbt!]
    \centering
    \includegraphics[width=\linewidth]{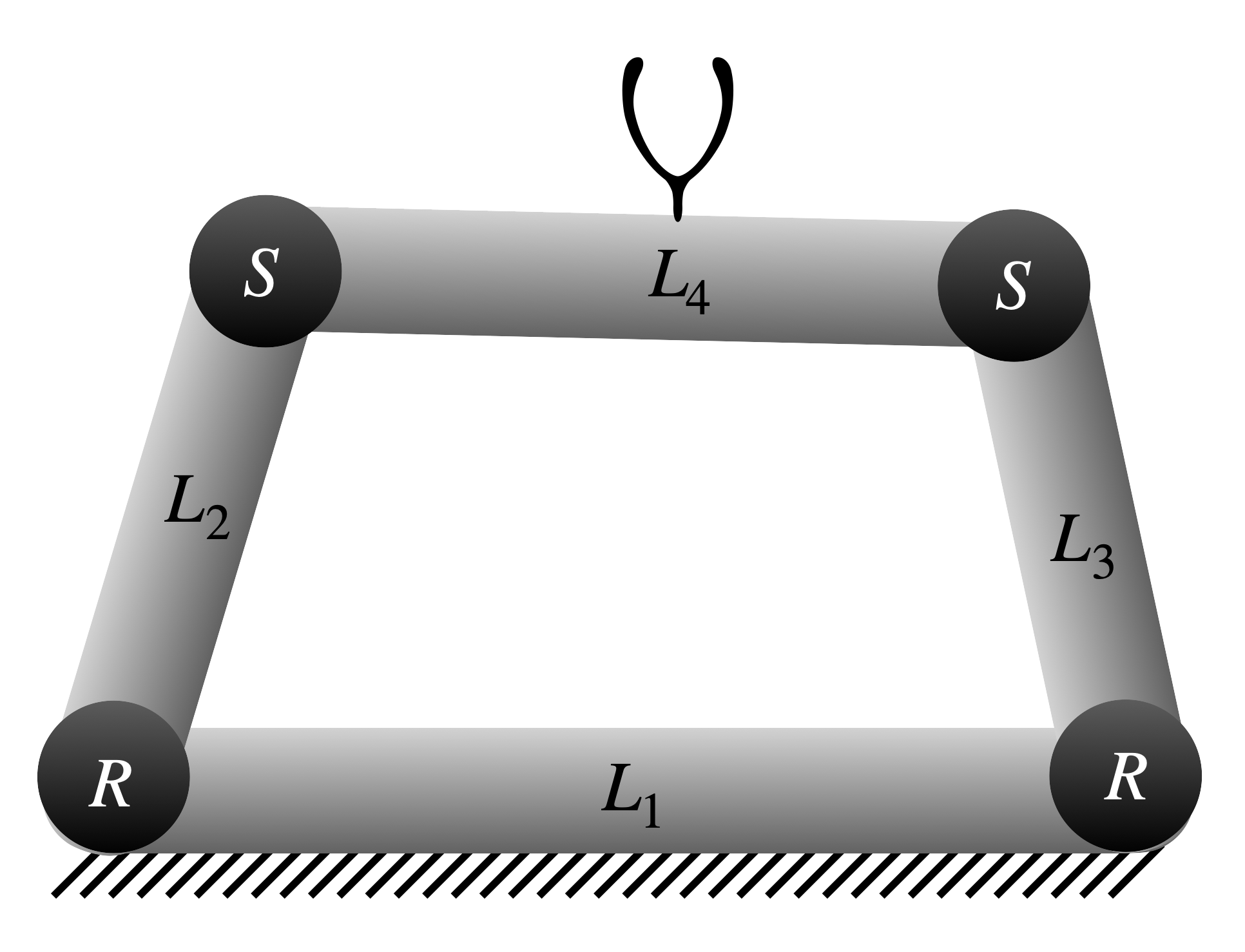}
    \caption{Schematic diagram of another manipulator having its end-effector uncontrollable by its actuator (an example for condition 8).}
    \label{fig_illustration_9}
  \end{figure}

If it is the case that all the joints happen to be connected to one link of a part, then it may be either a case containing superfluous DOF or an invalid case. If both the base link and the end-effector link lie in the same part then it would be a case containing superfluous DOF, and if they lie in different parts then that would be an invalid case because the angular velocity of the part that contains the end-effector about the axis passing through the two spherical joints cannot be controlled due to the fact that only revolute and prismatic joints are considered to be capable of being actuators in this study. An example of this case can be seen in Figure \ref{fig_illustration_9}.

The coupling matrix for the split of $p_1=\{1,2,3\}$ and $p_2=\{4\}$ is shown in Equation \eqref{eq:coupling_matrix5}. Even though all the spherical joints are connected to the same link $L_4$, the base link $L_1$ is in $p_1$ whilst the end-effector link $L_4$ is in $p_2$. Hence such matrices are considered to be invalid in this study.

\begin{equation}
\label{eq:embedded_matrix5}
    \widetilde{M} = \begin{pNiceMatrix}[margin]
    \tikzmark{c1_1}{L_1} & R & R & \Block[draw,fill=blue!15,rounded-corners]{3-1}{} O \\
    R & \tikzmark{c1_2}{L_2} & O & S \\
    R & O & \tikzmark{c1_3}{L_3} & S \\
    O & S & S & \tikzmark{c2_1}{L_4}
    \end{pNiceMatrix}
\end{equation}

\begin{tikzpicture}[overlay,remember picture]
     \draw[opacity=.2,line width=3mm,line cap=round] (c1_1.center) -- (c1_3.center);
     \draw[opacity=.2,line width=3mm,line cap=round] (c2_1.center) -- (c2_1.center);
\end{tikzpicture}
\begin{equation}
\label{eq:coupling_matrix5}
    C = \begin{bmatrix}
    O \\
    S \\
    S
    \end{bmatrix}
\end{equation}

In case of superfluous DOF, the count of superfluous DOF for the matrix is incremented by 1, so that the final number of superfluous DOF can be removed from the DOF computed using Kutzbach criterion for a more accurate calculation of DOF. The condition for checking for invalid matrix in this case is the eighth condition imposed in this study.

All these conditions (from 4 to 8) are concisely shown in Algorithm \ref{alg:conditions45678910}.

\begin{algorithm}
\caption{Pseudocode for conditions 4, 5, 6, 7 and 8}\label{alg:conditions45678910}
\begin{algorithmic}[1]
\Require Adjacency matrix $M$.\;
\Ensure The boolean values $c_4, c_5, c_6, c_7, c_8$ and $D_e$.\;
\State Assume $c_4, c_5, c_6, c_7$ and $c_8$ are true by default.\;
\State Set superfluous DOF $s = 0$.\;
\State Get all combinations of two parts of manipulator as $A$.
\For{$i$ through $A$}
\State $p_1$ = Part 1 of $i$.\;
\State $p_2$ = Part 2 of $i$.\;
\State $p_{12}$ = Joined list of $p_1$ and $p_2$.\;
\State Get the coupling matrix of these two parts as $C$.\;
\State $j = $ number of joints in $C$.\;
\If{j == 1}
\If{Both the base link and the end-effector link are in the same part}
\State $c_4=false$.\;
\State Break the loop.\;
\ElsIf{The joint in $C$ is either cylindrical or spherical}
\State $c_5=false$.\;
\State Break the loop.\;
\EndIf
\ElsIf{j>=2}
\If{all the joints in $C$ are connected to only one link of a part}
\If{both the base link and the end-effector link lie in the same part}
\State $c_6=false$.\;
\State Break the loop.\;
\EndIf
\EndIf
\If{j == 2}
\If{all the joints in $C$ are spherical joints}
\If{both the joints in $C$ are connected to only one link of a part}
\If{base link is in one part and end-effector link is in another part}
\State $c_8=false$.\;
\State Break the loop.\;
\Else
\State $s = s + 1$.\;
\EndIf
\Else
\State $c_7=false$.\;
\State Break the loop.\;
\EndIf
\EndIf
\EndIf
\EndIf
\EndFor
\State $D = $ DOF of $M$ based on Kutzbach criterion.\;
\State $D_e=D-s$.\;
\State Return $c_4, c_5, c_6, c_7, c_8$ and $D_e$.\;
\end{algorithmic}
\end{algorithm}

\subsection{Conditions 9, 10, 11 and 12}

The ninth, tenth, eleventh and twelfth conditions are grouped together as they all use the new effective DOF $D_n=D_k-s$, where $D_k$ is the DOF computed using Kutzbach criterion and $s$ is the number of superfluous DOF. The ninth condition is that the sum of the revolute and prismatic joints should be greater than or equal to $D_n$. The tenth condition imposed is that $D_n$ should be greater than or equal to 1.

The eleventh condition that is imposed is that each path that connects the base link and the end-effector link should together be able to facilitate independent components of motion greater than or equal to the effective degree of freedom $D_n$ of the manipulator. For example, the adjacency matrix corresponding to the planar manipulator shown in Figure \ref{fig_illustration_3} is shown in Equation \eqref{eq:main_matrix6}. There exist two paths that connect the base link and the end-effector link, which are shown in Equation \eqref{eq:embedded_matrix8} and Equation \eqref{eq:embedded_matrix9}. The two paths are\newline

\qquad Path 1: $L_1-R-L_2-R-L_3-R-L_4-R-L_5$

\qquad Path 2: $L_1-R-L_5$\newline

Even though Gruebler's criterion gives the DOF to be 2, the second path has only one revolute joint and therefore can only facilitate one revolute motion which is less than the expected DOF. Hence, due to this constraint, the end-effector would eventually have only 1 DOF. The other DOF is lost somewhere in Path 1. Such matrices are considered to be invalid in this study. Finally, the twelfth condition is to remove redundant manipulators. Since the 3D space can accommodate only 6 independent DOF (three linear and three angular components of velocities of the end-effector) to the end-effector, the matrices that have more than 6 DOF are considered invalid in this study. Additionally, since prismatic joints cannot contribute to revolute motion, if any path from the base link to the end-effector link happens to be having only prismatic joints, then the DOF should be at most 3 since the 3D space can accommodate only three independent linear components of end-effector's velocity. A pseudocode for these conditions is shown in Algorithm \ref{alg:condition_11_and_12}.

\begin{equation}
\label{eq:main_matrix6}
M=\left[\begin{matrix}L_{1} & R & O & O & R \\
R&L_{2}&R&O&O\\
O&R&L_{3}&R&O\\
O&O&R&L_{4}&R\\
R&O&O&R&L_{5}
\end{matrix}\right]
\end{equation}

\begin{equation}
\label{eq:embedded_matrix8}
    M = \begin{pNiceMatrix}[margin]
    \tikzmark{c1_1}{L_1} & \tikzmark{c1_2}{R} & O & O & R \\
R&\tikzmark{c1_3}{L_2}&\tikzmark{c1_4}{R}&O&O\\
O&R&\tikzmark{c1_5}{L_3}&\tikzmark{c1_6}{R}&O\\
O&O&R&\tikzmark{c1_7}{L_4}&\tikzmark{c1_8}{R}\\
R&O&O&R&\tikzmark{c1_9}{L_5}
    \end{pNiceMatrix}
\end{equation}

\begin{tikzpicture}[overlay,remember picture]
     \draw[opacity=.2,line width=3mm,line cap=round, color=red] (c1_1.center) -- (c1_2.center);
     \draw[opacity=.2,line width=3mm,line cap=round, color=red] (c1_2.center) -- (c1_3.center);
     \draw[opacity=.2,line width=3mm,line cap=round, color=red] (c1_3.center) -- (c1_4.center);
     \draw[opacity=.2,line width=3mm,line cap=round, color=red] (c1_4.center) -- (c1_5.center);
     \draw[opacity=.2,line width=3mm,line cap=round, color=red] (c1_5.center) -- (c1_6.center);
     \draw[opacity=.2,line width=3mm,line cap=round, color=red] (c1_6.center) -- (c1_7.center);
     \draw[opacity=.2,line width=3mm,line cap=round, color=red] (c1_7.center) -- (c1_8.center);
     \draw[opacity=.2,line width=3mm,line cap=round, color=red] (c1_8.center) -- (c1_9.center);
\end{tikzpicture}

\begin{equation}
\label{eq:embedded_matrix9}
    M = \begin{pNiceMatrix}[margin]
    \tikzmark{c3_1}{L_1} & R & O & O & \tikzmark{c3_2}{R} \\
R&L_2&R&O&O\\
O&R&L_3&R&O\\
O&O&R&L_4&R\\
R&O&O&R&\tikzmark{c3_3}{L_5}
    \end{pNiceMatrix}
\end{equation}

\begin{tikzpicture}[overlay,remember picture]
     \draw[opacity=.2,line width=3mm,line cap=round, color=red] (c3_1.center) -- (c3_2.center);
     \draw[opacity=.2,line width=3mm,line cap=round, color=red] (c3_2.center) -- (c3_3.center);
\end{tikzpicture}

\begin{algorithm}
\caption{Pseudocode for conditions 9, 10, 11 and 12}\label{alg:condition_11_and_12}
\begin{algorithmic}[1]
\Require Adjacency matrix $M$ and effective DOF $D_e$.\;
\Ensure The boolean values $c_{9}, c_{10}, c_{11}$ and $c_{12}$.\;
\State Assume $c_{9}, c_{10}, c_{11}$ and $c_{12}$ are true by default.\;
\If{$n_r+n_p<D_e$}
\State $c_9=false$\;
\ElsIf{$D_e<1$}
\State $c_{10}=false$\;
\EndIf
\State Get all paths that connect the base-link and the end-effector link of the manipulator, as $P$.
\For{each path $i$ through $P$}
\State Get all the joints in path $i$, as $J$.\;
\State $m_r = $ number of revolute joints in $J$.\;
\State $m_p = $ number of prismatic joints in $J$.\;
\State $m_c = $ number of cylindrical joints in $J$.\;
\State $m_s = $ number of spherical joints in $J$.\;
\State $m_t = m_r+m_p+2m_c+3m_s$.\;
\If{$m_t<D_e$}
\State $c_{11} = false$.\;
\State Break the loop.\;
\EndIf
\If{$m_t==m_p$}
\If{$D_e>=4$}
\State $c_{12} = false$.\;
\State Break the loop.\;
\EndIf
\Else
\If{$D_e>=7$}
\State $c_{12} = false$.\;
\State Break the loop.\;
\EndIf
\EndIf
\EndFor
\State Return $c_{9}, c_{10}, c_{11}$ and $c_{12}$.\;
\end{algorithmic}
\end{algorithm}

Finally, for each matrix, each of these 11 conditions is checked and if any condition is not satisfied then that matrix is removed from the list. A pseudocode for the same is shown in Algorithm \ref{alg:invalid}.

\begin{algorithm}
\caption{Pseudocode for removing invalid matrices}\label{alg:invalid}
\begin{algorithmic}[1]
\Require List of all enumerated adjacency matrices $A$.\;
\Ensure The list $A$ is reduced to a list that does not consist of invalid matrices.\;
\State Initialise $i=0$.\;
\While{$i$ is less than the length of list $A$}
\State $M=$ the $i$-th element of list $A$.
\State Assign the invalid flag $F$ as $true$.\;
\State Get $c_1$ and $c_2$ for $M$.\;
\If{$c_1==true$ and $c_2==true$}
\State Get $c_3$ for $M$.\;
\If{$c_3==true$}
\State Get $c_4, c_5, c_6, c_7, c_8$ and $D$ for $M$.\;
\If{$c_4==true$ and $c_5==true$ and $c_6==true$ and $c_7==true$ and $c_8==true$}
\State Get $c_9, c_{10}, c_{11}$ and $c_{12}$ for $M$ by using effective DOF $D_e$.\;
\If{$c_9==true$ and $c_{10}==true$ and $c_{11}==true$ and $c_{12}==true$}
\State $F=false$.\;
\State $i=i+1$.\;
\EndIf
\EndIf
\EndIf
\EndIf
\If{$F==true$}
\State Remove the $i$-th matrix from list $A$.\;
\EndIf
\EndWhile
\end{algorithmic}
\end{algorithm}

\subsection{Isomorphism detection and elimination}

Once the invalid matrices are eliminated from the list, the list would then be required to undergo isomorphism detection and elimination. The method permutes all the links of an adjacency matrix except the base and the end-effector links, by preserving the joint- connections between the links in each permuted item. The list of permuted matrices thus formed, amounts to the exhaustive set of possible representations of the topology of the robot. Other than one among this list, there is no other way to represent the topology of the robot using adjacency matrix. Hence each matrix from the permuted list is compared with other adjacency matrices to check if a match could be found. If no match is found then this adjacency matrix is not isomorphic with any other enumerated adjacency matrix. A pseudocode for the same is shown in Algorithm \ref{alg:isomorphism}.

\begin{algorithm}
\caption{Algorithm for elimination of isomorphic matrices}\label{alg:isomorphism}
\begin{algorithmic}[1]
\Require List $A$ of valid adjacency matrices.
\Ensure The list $A$ is reduced to a list of completely non-isomorphic adjacency matrices.
\State Initialise $i=0$.\;
\While{$i$ is less than the length of $A$}
\State Get $i$-th adjacency matrix of $A$ as $M$.\;
\State Get the list of all possible isomorphic adjacency matrices of $M$ as $I$.\;
\State Get the list of matrices of $A$ from $i+1$-th index to the last index, as $B$.\;
\State Initialise isomorphism identification flag $F = false$.\;
\For{$j$ through $I$}
\If{$j$ is found in $B$}
\State $F = true$.\;
\State Break the loop.\;
\EndIf
\EndFor
\If{$F == false$}
\State Remove $M$ from $A$.
\Else
\State continue.\;
\EndIf
\EndWhile
\end{algorithmic}
\end{algorithm}

\section{Results}

In the enumeration of four-link manipulators, the total number of posible adjacency matrices were 15625, which after elimination of invalid matrices had been reduced to 208, and further reduced to 104 after elimination. Among 104, 96 are of 1-DOF and 8 are of 3-DOF.

In the enumeration of five-link manipulators, the total number of possible adjacency matrices were 9765625, and after elimination of invalid matrices the list was reduced to 14829, and furthermore after elimination of isomorphism it was reduced to 2537, out of which 1826 are of 1-DOF, 696 are of 2-DOF and 15 are of 4-DOF. The counts of retained matrices after each set of conditions is passed, are shown in Table \ref{Tab:numbers_of_invalid_and_isomorphic}.

\begin{table}[h] \caption{The number of retained matrices after each set of conditions is passed} \centering \begin{tabular}{|l|l|l|l|}
    \hline
    \textbf{Conditions} & \textbf{3 links} & \textbf{4 links} & \textbf{5 links} \\ 
    \hline
    All possible permutations & $125$ & $15625$ & $9765625$ \\
    Conditions 1 and 2 & $48$ & $3412$ & $409640$ \\
    Condition 3 & $42$ & $1640$ & $116865$ \\
    Conditions 4, 5, 6, 7 \text{and} 8 & $6$ & $346$ & $25781$ \\
    Conditions 9, 10, 11 \text{and} 12 & $4$ & $208$ & $15018$ \\
    Isomorphism elimination & $4$ & $104$ & $2537$ \\
\hline\end{tabular} \label{Tab:numbers_of_invalid_and_isomorphic} \end{table}

\subsection{Enumeration of manipulators of DOF 1}

To reduce the complexity, four-link manipulators of 1DOF are enumerated.

\begin{table}[h] \caption{Classes of 1-DOF Manipulators of three links} \centering \begin{tabular}{|l|l|l|}
    \hline
    \textbf{Class} & \textbf{Type} & \textbf{Count} \\ 
    \hline
    Class 1 & $R^2CS$ & $18$ \\
    Class 2 & $R^2S^2$ & $3$ \\
    Class 3 & $PPCS$ & $36$ \\
    Class 4 & $RPS^2$ & $6$ \\
    Class 5 & $RC^3$ & $6$ \\
    Class 6 & $P^2CS$ & $18$ \\
    Class 7 & $P^2S^2$ & $3$ \\
    Class 8 & $PC^3$ & $6$ \\
    \hline
    \textbf{Total} &  & $\textbf{96}$ \\
\hline\end{tabular} \label{Tab:table_dof1cl_1} \end{table}
After permuting the off-diagonal elements of adjacency matrix with all possible values and eliminating invalid and isomorphic adjacency matrices, 96 distinct adjacency matrices are finally obtained. Based on the sets of joints involved, the enumerated manipulators are classified into 8 classes. The description of each class along with the count of manipulators is shown in table \ref{Tab:table_dof1cl_1}. Schematic representations of manipulators of all the 8 classes can be found in a GitHub repository \cite{fivelinkenumeration} shown in the references, in which the first link is the base link and the dark circle inscribed with `E' represents the end-effector point.

\subsection{Enumeration of manipulators of DOF 2}

Enumeration of 2-DOF manipulators is done from the list of three-link, four-link and five-link manipulators.

\begin{table}[h] \caption{Classes of 2-DOF Manipulators amongst 3, 4 and 5 links} \centering \begin{tabular}{|l|l|l|}
    \hline
    \textbf{Class} & \textbf{Type} & \textbf{Count} \\ 
    \hline
    Class 1 & $R^{3}CS$ & $64$ \\
    Class 2 & $R^{3}S^{2}$ & $9$ \\
    Class 3 & $R^{2}PCS$ & $192$ \\
    Class 4 & $R^{2}PS^{2}$ & $27$ \\
    Class 5 & $R^{2}C^{3}$ & $28$ \\
    Class 6 & $RP^{2}CS$ & $192$ \\
    Class 7 & $RP^{2}S^{2}$ & $27$ \\
    Class 8 & $RPC^{3}$ & $56$ \\
    Class 9 & $P^{3}CS$ & $64$ \\
    Class 10 & $P^{3}S^{2}$ & $9$ \\
    Class 11 & $P^{2}C^{3}$ & $28$ \\
    Class 12 & $R^2$ & $1$ \\
    Class 13 & $RP$ & $2$ \\
    Class 14 & $P^2$ & $1$ \\
    \hline
    \textbf{Total} &  & $\textbf{700}$ \\
\hline\end{tabular} \label{Tab:table_dof1cl_2} \end{table}

After permuting the off-diagonal elements of adjacency matrix with all possible values and eliminating invalid and isomorphic adjacency matrices, 700 distinct adjacency matrices are finally obtained. Based on the sets of joints involved, the enumerated manipulators are classified into 14 classes. The description of each class along with the count of manipulators is shown in table \ref{Tab:table_dof1cl_2}. Schematic representations of manipulators of all the 14 classes can be found in the GitHub repository \cite{fivelinkenumeration} shown in the references, in which the first link is the base link and the dark circle inscribed with `E' represents the end-effector point.

\subsection{Enumeration of manipulators of DOF 3}
Enumeration of 3-DOF manipulators is done from the list of three-link, four-link and five-link manipulators. 

\begin{table}[h] \caption{Classes of 3-DOF Manipulators amongst 3, 4 and 5 links} \centering \begin{tabular}{|l|l|l|}
    \hline
    \textbf{Class} & \textbf{Type} & \textbf{Count} \\ 
    \hline
    Class 1 & $R^3$ & $1$ \\
    Class 2 & $R^2P$ & $3$ \\
    Class 3 & $RP^2$ & $3$ \\
    Class 4 & $P^3$ & $1$ \\
    \hline
    \textbf{Total} &  & $\textbf{8}$ \\
\hline\end{tabular}  \label{Tab:table_dof1cl_3} \end{table}

After permuting the off-diagonal elements of adjacency matrix with all possible values and eliminating invalid and isomorphic adjacency matrices, 8 distinct adjacency matrices are finally obtained. Based on the sets of joints involved, the enumerated manipulators are classified into 4 classes. The description of each class along with the count of manipulators is shown in table \ref{Tab:table_dof1cl_3}. Schematic representations of manipulators of all the 4 classes can be found in the GitHub repository \cite{fivelinkenumeration} shown in the references, in which the first link is the base link and the dark circle inscribed with `E' represents the end-effector point.

\subsection{Enumeration of manipulators of DOF 4}
Enumeration of 4-DOF manipulators is done from the list of three-link, four-link and five-link manipulators. 

\begin{table}[h] \caption{Classes of 4-DOF Manipulators amongst 3, 4 and 5 links} \centering \begin{tabular}{|l|l|l|}
    \hline
    \textbf{Class} & \textbf{Type} & \textbf{Count} \\ 
    \hline
    Class 1 & $R^4$ & $1$ \\
    Class 2 & $R^3P$ & $4$ \\
    Class 3 & $R^2P^2$ & $6$ \\
    Class 4 & $RP^3$ & $4$ \\
    \hline
    \textbf{Total} &  & $\textbf{15}$ \\
\hline\end{tabular}  \label{Tab:table_dof1cl_4} \end{table}

After permuting the off-diagonal elements of adjacency matrix with all possible values and eliminating invalid and isomorphic adjacency matrices, 15 distinct adjacency matrices are finally obtained. Based on the sets of joints involved, the enumerated manipulators are classified into 4 classes. The description of each class along with the count of manipulators is shown in table \ref{Tab:table_dof1cl_4}. Schematic representations of manipulators of all the 4 classes can be found in the GitHub repository \cite{fivelinkenumeration} shown in the references, in which the first link is the base link and the dark circle inscribed with `E' represents the end-effector point.

\section{Limitations}
The criterion used for calculating DOF in this study is limited to Kutzbach criterion. But there are some valid mechanisms that are classified as structures by Kutzbach criterion. An example of this is four-bar PPPP spatial mechanism. The reason for this is that since all the motion allowed/facilitated by any serial/parallel combination of prismatic joints is only linear but not angular, the Jacobian has non-zero elements only for half of its rows (the upper three rows), and any three linearly independent columns of Jacobian would reach the threshold of a structure and any fourth column would enable reciprocation with the set of other three joints, thereby making the motion possible. On the other hand, if there were other kinds of joints existing in the closed-loop mechanism that facilitate rotary motion as well, then all the six rows of Jacobian could be non-zero, and hence six linearly independent columns of Jacobian would reach the threshold of a structure this time, requiring a 7th column to enable reciprocation with those six linearly independent columns in order to make motion possible in a general case. This is the reason why four-bar RRRR spatial kinematic chain does not move whilst four-bar PPPP spatial kinematic chain does, and 7-bar RRRRRRR mechanism is the minimum analogous mechanism that allows motion. But since Kutzbach criterion does not distinguish prismatic joint from revolute joint but considers them to be equivalently contributing, this criterion is not sufficient to determine the exact DOF of mechanisms in some cases. Some of the criteria shown in sub-section \ref{criteria} are developed by analysing the Jacobian of each manipulator after enumerating all the possible adjacency matrices, and hence there exist some manipulators in the enumerated list that do not satisfy the Kutzbach criterion yet have the required DOF. Hence the enumeration presented in this study is not complete. The completeness of enumeration is suggested as future scope.

\section{Conclusion}

The concept of adjacency matrix is used for enumeration of spatial manipulators with four types of joints, namely revolute, prismatic, cylindrical and spherical joints, by analysing its applicability to spatial manipulators. Criteria for eliminating invalid and isomorphic adjacency matrices are presented. Finally, 96 1-DOF manipulators of 8 classes are enumerated from 4-link adjacency matrices, 700 2-DOF manipulators of 14 classes, 8 3-DOF manipulators of 4 classes and 15 4-DOF manipulators of 4 classes are enumerated from 3-link, 4-link and 5-link adjacency matrices. The schematic diagrams of the manipulators are presented. This set of enumerated manipulators is used in the companion study \cite{synthesis_unpublished} on dimensional synthesis wherein dimensions are presented for each manipulator for optimal performance in a particular context that is described in the companion paper. The two studies together aim to provide to the designer an atlas of manipulators along with their optimal dimensions and their ranking, which can be useful to the designer to choose the best manipulator relevant to the context mentioned in the companion paper.

Since the steps provided in this study seem to be sufficient for enumeration of spatial manipulators, this enumeration study is concluded with the presented manipulators to move on to the companion study of dimensional synthesis, although the enumeration study can be extended to higher number of links and with inclusion of more types of joints such as universal and helical joints. The authors suggest this as future scope.

\end{document}